\documentclass{article}
\usepackage{spconf,amsmath,graphicx}
\usepackage{subcaption}
\usepackage{booktabs}
\usepackage{subfiles}
\usepackage{floatrow}
\newfloatcommand{capbtabbox}{table}[][\FBwidth]
\usepackage{varwidth}
\usepackage[font=small,labelfont=bf,tableposition=top]{caption}
\DeclareCaptionLabelFormat{andtable}{#1~#2  \&  \tablename~\thetable}
\usepackage{color}
\usepackage{framed}
\usepackage{multicol}
\usepackage{comment}
\usepackage{float}
\usepackage{mathtools,amssymb}
\definecolor{shadecolor}{gray}{0.875}

\usepackage[T1]{fontenc}    % use 8-bit T1 fonts
\usepackage{hyperref}       % hyperlinks
\usepackage{url}            % simple URL typesetting
\usepackage{booktabs}       % professional-quality tables
\usepackage{amsfonts}       % blackboard math symbols
\usepackage{nicefrac}       % compact symbols for 1/2, etc.
\usepackage{microtype}      % microtypography

\usepackage{mathtools,amssymb}
\usepackage{algorithm}
\usepackage[noend]{algpseudocode}
\usepackage{algpseudocode}
\title{Pruning at a Glance: \\Global Neural Pruning for Model Compression}

\newcommand*{\affaddr}[1]{#1} 
\newcommand*{\affmark}[1][*]{\textsuperscript{#1}}

\name{Abdullah Salama\affmark[* $\mathparagraph$], Oleksiy Ostapenko\affmark[* $\mathsection$,], Tassilo Klein\affmark[*], Moin Nabi\affmark[*]}

\address{\affaddr{\affmark[*]SAP ML Research},
\affaddr{\affmark[$\mathparagraph$]Hamburg University of Technology},
\affaddr{\affmark[$\mathsection$]Humboldt University of Berlin}}

\begin{document}
\maketitle
\begin{abstract}
Deep Learning models have become the dominant approach
in several areas due to their high performance. Unfortunately,
the size and hence computational requirements
of operating such models can be considerably high. Therefore,
this constitutes a limitation for deployment on memory
and battery constrained devices such as mobile phones or
embedded systems. To address these limitations, we propose
a novel and simple pruning method that compresses
neural networks by removing entire filters and neurons according
to a global threshold across the network without
any pre-calculation of layer sensitivity. The resulting model
is compact, non-sparse, with the same accuracy as the non-compressed
model, and most importantly requires no special
infrastructure for deployment. We prove the viability
of our method by producing highly compressed models, namely VGG-16, ResNet-56, and ResNet-110 respectively on CIFAR10 without losing any performance
compared to the baseline, as well as ResNet-34 and ResNet-50 on
ImageNet without a significant loss of accuracy. We also provide a well-retrained 30\% compressed ResNet-50 that slightly surpasses the base model accuracy. Additionally,
compressing more than 56\% and 97\% of AlexNet
and LeNet-5 respectively. Interestingly, the resulted models' pruning patterns are highly similar to the other methods using
layer sensitivity pre-calculation step. Our method does
not only exhibit good performance but what is more also
easy to implement.
\end{abstract}
\section{Introduction}
While deep learning models have become the method of choice for a multitude of applications, their training entails optimizing a large number of parameters at extensive computational cost (energy, memory footprint, inference time).

This effectively limits their deployment on storage and battery constrained devices, such as mobile phones and embedded systems. To study their parameterization behavior, \cite{Redundancy} revealed significant parameters' redundancy in several deep learning models. To reduce this redundancy and compress deep learning models without loss in accuracy several approaches have been proposed.~\cite{optbraindmg,optbrainsgn} proposed pruning weights by optimizing network complexity using second-order derivative information. However, due to the high computational overhead of second order derivatives,~\cite{approx1,approx2} explored low-rank approximations to reduce the size of the weight tensors.

\begin{figure}[h!]
\begin{center}                      
   \includegraphics[width=0.8\linewidth]{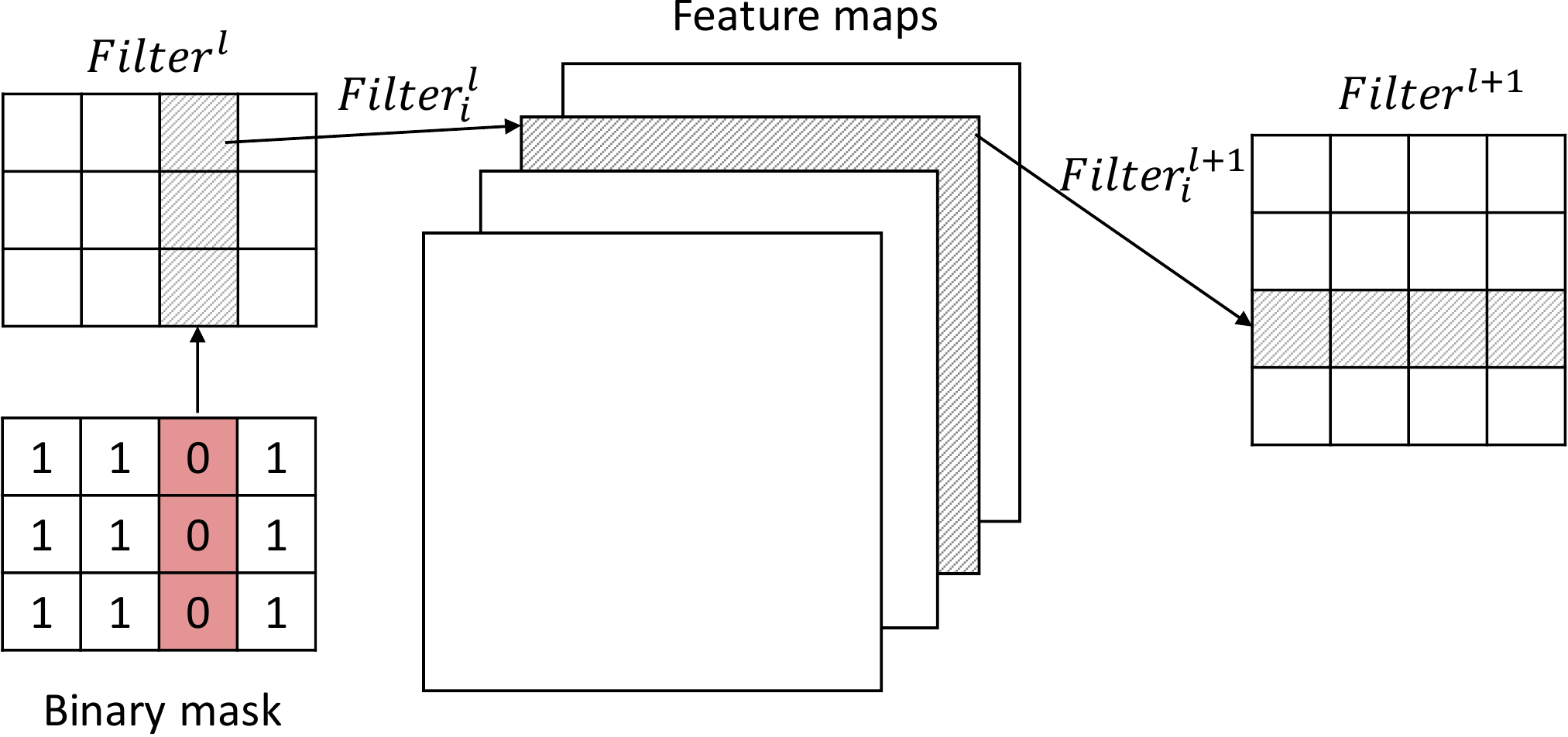}
\end{center}
   \caption{Filter pruning for each layer; $Filter^{l}$ refers to all filters in layer $l$; $Filter_i^l$ corresponds to filter $i$ in a certain layer $l$. When a filter is removed from a layer, the corresponding input with dependencies to the next layer is pruned as well. Binary mask is used to zero-out a certain filter to be later removed.}
\label{fig:teaser}
\end{figure}

Another line of work \cite{Pruning1, trainspnn}, proposed to prune individual layer weights with the lowest absolute value (non-structural sparsification of layer weights).
\cite{Pruning2} followed the same strategy, additionally incorporating quantization and Huffman coding to further boost compression.
While the aforementioned methods considered every layer independently,
\cite{class_blind_pruning} pruned the network weights according to a global threshold for all the network, e.g. individual layer weights are pruned according to their magnitude as compared to all weights in the network. Weight pruning is referred to as non-structured pruning as removing a part of network structure is not essentially guaranteed because of the resulted unstructured sparsity.

\begin{figure*}[]
  \centering           
  \centerline{\includegraphics[width=\linewidth]{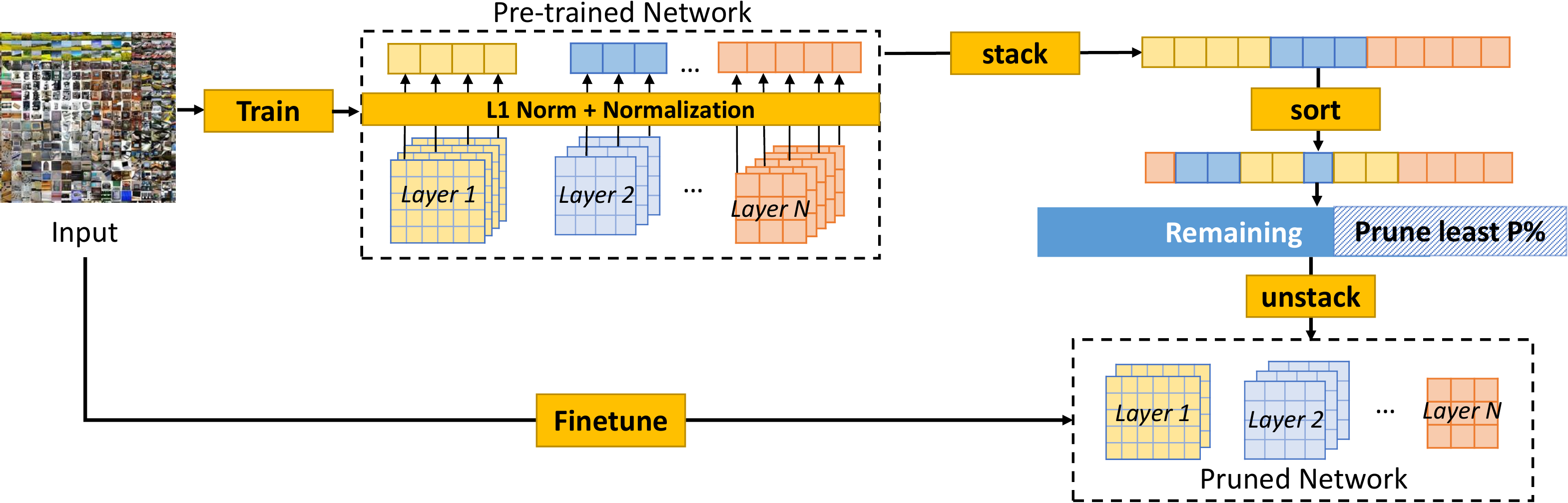}}
\caption{Overview of proposed method: as the first step, the L1-norm of each filter in each layer is calculated, then the calculated L1-norm of each filter is normalized according to its number of kernel weights, followed by stacking all normalized norms of all filters from all layers. As a last step, we perform sorting and pruning procedures of the filters corresponding to the least $p\%$ of normalized norms.}
\label{fig:overview}
\end{figure*}

Noteworthy, all non-structured pruning approaches, generally result in high sparsity models that require special hardware and software. Structured pruning alleviates this by removing whole filters or neurons, producing a non-sparse compressed model. In this regard, \cite{pruningfilters} proposed channel-wise pruning according to the L1-norm of the corresponding filter. \cite{learnssinnn} learned a compact model based on learning structured sparsity of different parameters. An algorithm was implemented to remove redundant neurons iteratively on fully connected layers in \cite{dfpruning}. In \cite{ddpruning}, connections leading to weak activations were pruned. Finally, \cite{NISP_paper} pruned the least important neurons after measuring their importance with respect to the penultimate layer. Generally, in past work, each layer's importance/sensitivity to pruning was evaluated separately and each layer was pruned accordingly. This work features two key components:
\begin{itemize}
\item[(a)] \textbf{Global Pruning}: All layers are considered simultaneously for pruning; considering layers simultaneously was first introduced by \cite{class_blind_pruning} to prune individual weights and was referred as class-blind pruning.
\item[(b)] \textbf{Structured Pruning}: Removal of entire filters or neurons instead of individual weights. Fig.~\ref{fig:teaser} shows the schematic of proposed structured pruning method for a single layer.
\end{itemize}
To the best of our knowledge, we are the first to use these two components together to consider structured pruning across all layers simultaneously. This is achieved by pruning filters based on their relative L1-norm compared to the sum of all filters' L1-norms across the network, instead of pruning filters according to their L1-norm within the layer \cite{pruningfilters}, inducing a global importance score for each filter (Fig \ref{fig:overview}). Most importantly, due to the global importance score derived, we do not impose any restrictions on which layers or filters to be pruned. This is in contrast to the limitations of \cite{pruningfilters}, which propose pre-calculation of layers' sensitivity to pruning and consequently avoid pruning sensitive layers completely, assuming that a layer containing some high sensitive filters is inherently very sensitive. Such assumption is not accurate, as each layer can contain different filter sensitivities and subsequently least sensitive filters can be pruned. In our method, each filter's relative L1-norm represents its true sensitivity to being pruned. The contributions of this paper are two-fold:
\begin{itemize}
    \item[\textbf{i)}]  Proposing a structured global pruning technique to compress the network by removing whole filters and neurons, which results in a compact non-sparse network with the same baseline performance.
    \item[\textbf{ii)}] Introducing a visualization of global filter importance to devise the pruning percentage of each layer.
\end{itemize}

As a result, the proposed approach achieves high compression percentages on VGG-16, ResNet-56 and ResNet-110 on the CIFAR10 dataset \cite{krizhevsky2014cifar}. As well as high compression results of AlexNet, ResNet-34 and ResNet-50 on ImageNet ILSVRC-2012 dataset \cite{ImageNet_Paper}.

\label{sec:intro}
\section{Background}
\label{gen_inst}

Several work studied compressing deep learning models while maintaining the same baseline performance. The most related works to our method can be categorized as following:

\subsection{Weight Pruning} Recently, \cite{Pruning1} proposed pruning multiple deep learning models up to 92\% by zeroing out the least percentage of weights per layer based on a layer weight standard of deviation. A follow-up work by \cite{Pruning2} incorporated pruning by quantization and Huffman coding to limit the non-sparse weight representation, and compressing the resulting weight representation, respectively. Moreover, \cite{trainspnn} proposed binary masks that are learned with the weights simultaneously during training. Thereby, the weights are multiplied by the masks, resulting in a sparse parameter representation. While all these methods produce high sparsity models, their hyperparameter optimization is very complex. Most importantly, in order to benefit from high sparsity models, special hardware or software is required.

\subsection{Structured Pruning} To address the limitation of sparsity-induced pruning, structured pruning was introduced. The underlying idea is to produce structured sparsity, e.g. remove parts of the structure, equivalent to filters/neurons in CNNs. \cite{pruningfilters} pruned filters having the lowest weights in terms of L1-norm within each layer, eventually removing filters of a trained network, followed by retraining the network to regain accuracy. \cite{learnssinnn} proposed a method that learns a compact model based on learning structured sparsity of filters, channels, filter shapes, and layer depth of the base model. Moreover, \cite{dfpruning} implemented a data-free algorithm to remove redundant neuron by neuron iteratively on fully connected layers. Also, \cite{ddpruning} identified redundant connections based on analyzing weak neurons on a validation set according to their weak activations. Thus, pruning the connections leading to the weak neurons iteratively until a compact model is obtained. In \cite{DDSSSfDDN}, a sparsity regularizer is imposed on outputs of various network entities such as neurons, groups or residual blocks after introducing a scale factor to scale the structures' output. By minimizing the scaling factors, speed-up on some models was achieved as well as limited compression results. Finally, \cite{NISP_paper} pruned neurons by measuring the neuron importance with respect to the last layer before the classification layer.

Similar to \cite{pruningfilters}, our proposed approach prunes filters employing the L1-norm. However, instead of choosing which filter to prune according to its relative L1-norm within its layer, we prune according to the relative norm with respect to the all layers.
\section{Structured Global Pruning}
Consider a network with a convolutional (conv) layer and a fully connected (fc) layer. We denote each filter $Filter_i$, where $i \in [1,F]$, and $F$ is the total number of filters in the conv layer. Each filter is a 3D kernel space consisting of channels, where each channel is associated with 2D kernel weights. For the fc layer, we denote $W_{m}$, a 1-D feature space containing all the weights connected to certain neuron $Neuron_m$, with $m \in [1,N]$ and $N$ denoting the number of neurons. It should be noted that we do not prune the classification layer.

Each pruning iteration in our approach is structured as presented in Algorithm \ref{algorithm}. To explain the method in detail, we first calculate the L1-norm of each filter by calculating the L1-norm of all its weights, followed by normalizing each result by the number of weights. After that, all normalized L1-norms components from all layers are stacked together and sorted in an ascending order. Then, a global threshold is chosen according to the value corresponding to the least p\% of the stacked norms. Finally, each filter having a normalized norm less than the global threshold is removed. Similarly, for fully connected layers, norms for neurons are stacked with filters' norms. Moreover, a neuron is removed if the corresponding normalized L1-norm of its connected weights is less than the global threshold.
Next, we present important elements of our method as following:

\textbf{Importance calculation:}
Although pre-calculation of filters or layers' sensitivity to be pruned is not needed in our method, it can be visualized as part of the pruning criteria. In our algorithm, global pruning implies constructing a hidden importance score, which corresponds to the relative normalized L1-norm. For instance, the relevant importance for a certain filter in a conv layer with respect to the all other filters in all layers is the ratio between the filter's normalized norm and the sum of all filters' normalized norms across the network.

\textbf{Normalization:}
As each layer's filters have different number of kernel weights, we normalize filters' L1-norms by dividing each over the number of kernel weights corresponding to the filter (Line~\ref{norm1} and~\ref{norm2} as indicated in Algorithm \ref{algorithm}). Alternatively, without compensating for the number of weights, filters with more kernel weights would have higher probabilities of higher L1-norms, hence lower probability to get pruned.

\textbf{Retraining process:}
 Pruning without further adaption, results in accuracy loss. Therefore, in order to regain base performance, it is necessary for the model to be retrained. To this end, we apply an iterative pruning schedule that alternates between pruning and retraining. This is conducted until a maximum compression is reached without losing the base accuracy, e.g. the model is pruned with a certain percentage. Next the model is retrained, which is followed by an increase of the pruning percentage. This cycle is repeated until the termination criteria is met, i.e. the desired accuracy is obtained.
 
\begin{algorithm*}[h!]
\caption{Pruning  procedure}\label{algorithm}
\begin{algorithmic}[1]
\For{$i \gets 1$ to $F$}\Comment{loop over filters of a conv layer}
\State $L1\_conv(i) \gets sum(|Filter_i|)$ \Comment{calculate L1-norm of all channels' kernel weights of each filter}
\State $norm\_conv(i) \gets L1\_conv(i) / size(Filter_i)$ \label{norm1} \Comment{normalize by filter weights count}
\EndFor
\For{$m \gets 1$ to $N$}\Comment{loop over Neurons of a fc layer}
\State $L1\_fc(m) \gets sum(|W_m|)$ \Comment{for each Neuron, calculate L1-norm of incoming weights}
\State $norm\_fc(m) \gets L1\_fc(m) / size(W_m)$ \label{norm2} \Comment{normalize by number of weights connected}
\EndFor
\State $norms \gets stack (norm\_conv,norm\_fc)$ \Comment{stack all normalized norms from all layers}
\State $sorted \gets sort(norms)$ \Comment{sort ascendingly}
\State $threshold \gets perc(sorted,p)$ \Comment{threshold based on a percentage p of sorted norms values}
\For{$i \gets 1$ to $F$}
 \If{$norm\_conv_(i) < threshold$}
     \State $prune(Filter_i)$ \Comment{remove filter if its normalized norm is less than threshold}
 \EndIf
\EndFor
\For{$m \gets 1$ to $N$}
 \If{$norm\_fc(m) < threshold$}
     \State $prune(Neuron_m)$ \Comment{remove neuron if its normalized norm is less than threshold}
 \EndIf
\EndFor
\end{algorithmic}
\end{algorithm*}
\section{Experiments}
In order to assess the efficacy of the proposed method, the performance of our technique is evaluated on a set of different networks: LeNet-5 on MNIST~\cite{LeNetLecun}, a version of VGG-16~\cite{pruningfilters}, ResNet-56 and ResNet-110~(\cite{ResNetcite}) on CIFAR-10~\cite{krizhevsky2014cifar}, AlexNet~\cite{alexnetpaper}, ResNet-34 and ResNet-50 on ImageNet ILSVRC-2012~\cite{ImageNet_Paper}. Followed by analyzing some of the resulted pruning patterns. Then we analyze different components of our methods on LeNet, finally we analyze the effect of our method on the learning process.

\subsection{Experimental Settings}
We perform experimentation on three image classification datasets: MNIST \cite{LeNetLecun}, CIFAR-10~\cite{krizhevsky2014cifar} and ImageNet~\cite{ImageNet_Paper}. We asses the classification performance using some common CNN models: LeNet-5~\cite{LeNetLecun} for MNIST, an altered version of VGG-16~\cite{pruningfilters}, ResNet-56/110~\cite{ResNetcite} on CIFAR-10, AlexNet~\cite{alexnetpaper} and ResNet-34/50~\cite{ResNetcite} on ImageNet.

For our implementation, we use PyTorch \cite{Pytorchcite}. Also, AlexNet pretrained model~\cite{AlexNetpy_paper}, as well as ResNet models are obtained using the same framework. For pruning implementation, we use binary masks to represent a pruned parameter by zero and non-pruned by one. Binary masks zero-out the pruned parameters and restrict them from learning. Generally, when a filter is removed, the corresponding input channel in the next layer is removed, similarly if a neuron is removed, the corresponding weights connected to this neuron in the next layer is removed. Additionally, when a filter is pruned, the corresponding batch-normalization weight and bias applied to that filter are pruned accordingly.

\subsubsection{LeNet on MNIST}
First, we experiment with LeNet-5 on MNIST. LeNet-5 is a convolutional network that has two convolutional layers and two fully connected layers with total of 431K parameters. LeNet-5 has base error of 0.8\% on MNIST. Tab.~\ref{table:LeNettest} shows the compression statistics of the experiment.

\subsubsection{VGG-16 on CIFAR-10:}
VGG-16~\cite{VGG_Paper} was originally designed for the ImageNet dataset. \cite{VGGforCIFAR} applied an altered version of VGG-16 to achieve better classification performance on CIFAR-10. The model consists of 13 convolutional layers and 2 fully connected layers with a total of 15M parameters. Recently,~\cite{pruningfilters} proposed a modification incorporating Batch Normalization~\cite{BatchNorm_paper} after the convolutional layers and the first fully connected layer. 
For the training settings, we use identical settings as ResNet in~\cite{ResNetcite}, except after pruning where we retrain for 50 epochs and with an initial learning rate of 0.05.

\subsubsection{ResNet on CIFAR-10}
CIFAR-10 ResNet models have three stages of residual blocks for feature maps with sizes of $32\times32$, $16\times16$ and $8\times8$. Each stage has the same number of residual blocks. When the number of feature maps increases, the shortcut layer performs identity mapping, by padding extra zero entries with the increased dimension. As for training settings, we use identical training settings for ResNet-56/110 as~\cite{ResNetcite}, with the difference of having an initial learning rate of 0.05 during retraining for 50 epochs, which follows each pruning iteration.

\begin{table}[h!]
\begin{center}
\begin{tabular}{lllll}
\hline
  Network          & Model & AccGain\%       & Param\%        & \\ \hline
   VGG-16        & Li et al.-A \cite{pruningfilters}      & 0.15              & 64.00                   \\      
& Ours  & \textbf{0.16} & \textbf{86.10}         \\ \cline{2-5} 
           & Li et al.-A \cite{pruningfilters}      & 0.06              & 9.40                   \\      
ResNet-56           & Li et al.-B \cite{pruningfilters}      & 0.02             & 13.70                      \\              
 & NISP-56 \cite{NISP_paper}  & -0.03          & 42.60          &         \\
& Ours  & \textbf{0.08} & \textbf{47.86}         \\ \cline{2-5} 
           & Li et al.-A \cite{pruningfilters}      & 0.02              & 2.30                    \\      
 ResNet-110          & Li et al.-B \cite{pruningfilters}      & -0.23              & 32.40                \\   
  & NISP-110 \cite{NISP_paper}  &  -0.18             & 43.25                    \\
 & Ours  &       \textbf{0.03}          &  \textbf{53.06}              &         \\ \hline
\end{tabular}
\end{center}
\caption{Benchmark results on CIFAR10. AccGain\% denotes the accuracy improvement over the base accuracy; Param\% denotes the percentage of pruned parameters}
\label{table:Benchmark}
\end{table}

\begin{table}[h!]
\begin{center}
\begin{tabular}{lllll}
\hline
  Network          & Model & AccGain\%       & Param\%        & \\ \hline     
    & NISP-A\cite{NISP_paper}      & -1.43             & 33.77                     \\              
AlexNet        & NISP-D \cite{NISP_paper}  & \textbf{0.00}          & 47.09         &         \\
& Ours  & -1.10 & \textbf{56.17}         \\ \cline{2-5} 
           & ResNet-34-B \cite{pruningfilters}      & -1.06              & 10.8                    \\      
 ResNet-34          & NISP-34-A \cite{NISP_paper}  &  \textbf{-0.28}            & 27.14               \\   
  & NISP-34-B \cite{NISP_paper}  &  -0.92            & 43.68                    \\
 & Ours  &       -0.95          &  \textbf{45.25}              &         \\ \cline{2-5} 
       & NISP-50-A \cite{NISP_paper}  &  \textbf-0.21            & 27.12               \\   
      & NISP-50-B \cite{NISP_paper}  &  -0.89           & 43.82 \\
 ResNet-50 & Ours-minError  &       \textbf{0.09}          &  30          \\
  & Ours-minEpochs  &       -0.60          &  30      \\ 
    & Ours-maxCompr  &       -0.51          &  \textbf{45.65}         \\ \hline
\end{tabular}
\end{center}
\caption{Benchmark results on ImageNet. AccGain\% denotes the accuracy improvement over the base accuracy; Param\% denotes the percentage of pruned parameters}
\label{table:Benchmark}
\end{table}
\subsubsection{AlexNet on ImageNet}
We use the AlexNet PyTorch pre-trained model, which has 61 million parameters across 5 convolutional layers and 3 fully connected layers. The AlexNet PyTorch model achieves a top-1 accuracy of 56.55\%. After each pruning iteration, the network is retrained with an initial learning rate of 0.0005 (1/50 of the network's original initial learning rate) for 20 epochs. We use 5 pruning iterations for extracting the maximum pruning percentage.
\subsubsection{ResNet on ImageNet}
We use the ResNet-34 PyTorch pre-trained model, which has four stages of residual blocks for feature maps with sizes of $56\times56$, $28\times28$, $14 \times14$ and $7\times7$. The projection shortcut is used during feature maps' downsampling. The ResNet-34 and ResNet-50 PyTorch models achieve  top-1 accuracies of 73.3\% and 76.15\% respectively. After each pruning iteration, both networks are retrained with an initial learning rate of 0.005 (1/50 of the network's original initial learning rate). Due to the projection shortcut, we also prune the downsampling layers. As the identity shortcut and the last convolutional layer in the residual block have the same number of filters, we prune them equally. For ResNet-34, we use 5 pruning iterations, and retrain after each for 20 epochs to reach the maximum compressed model. While for ResNet-50 we use 4 pruning iterations, and retrain for 10 epochs, to explore the maximum pruning percentage that can be achieved with limited retraining and we denote this version of our method as "Ours-minEpochs". For ResNet-50, we change the number of retraining epochs to 10 and use 9 pruning iterations and then we retrain for a final step without increasing pruning. Using these setting, we produce two models "Ours-minError" that is aimed at having a minimum error or closest accuracy to the baseline and "Ours-maxCompr" that is aimed at having a high compression percentage.

\begin{table}[h!]
\begin{center}
\begin{tabular}{lccc}
\hline
Method   & AccGain\% & Param\% & Eff. Param\% \\ \hline
Han et al. \cite{Pruning1}  & 0.03      & 92.00    & 84.00              \\
Srinivas et al. \cite{trainspnn}  & -0.01      & 95.84    & 91.68             \\
Han et al. \cite{Pruning2}  & \textbf{0.06}     & \textbf{97.45}   & -              \\
Ours  & 0.05      & 97.40    & \textbf{97.40}              \\\hline
\end{tabular}
\end{center}
\caption{Results on LeNet-5. AccGain\% denotes the accuracy improvement over the base accuracy; Param.\% is the parameters' pruning percentage; Eff. Param\%. is the effective parameters' pruning percentage with taking into account the extra indices storage for non-structured pruning as studied by \cite{Sparsesaving}.} 
\label{table:LeNettest}
\end{table}

\subsection{Benchmark Results}
We report compression results on the existing benchmark of structured pruning~\cite{pruningfilters,NISP_paper} on CIFAR and ImageNet, and on MNIST with weight pruning methods~\cite{Pruning1,Pruning2, trainspnn}. As shown in Tab.~\ref{table:Benchmark}, we outperform the compression results reported by \cite{NISP_paper} on both ResNet-56 and ResNet-110 and on VGG-16 as reported by~\cite{pruningfilters}, both with a lower classification error even compared to the baseline. Additionally on ImageNet, we exceed other structured pruning methods in terms of network compression on AlexNet and highly compress ResNet-34 with a limited increase in the classification error. With using only 40 epochs in total during retraining, we exceed the compression results of one version of~\cite{NISP_paper} that used 90 epochs for retraining. Finally, using 100 epochs for retraining, we exceed the compression reported by~\cite{NISP_paper} without losing significant accuracy~("Ours-maxCompr") and produce another compressed model~("Ours-minError") that has a slightly better accuracy than the base model. 

To compare with other pruning methods on MNIST, we compare with non-structured pruning methods. In Tab. \ref{table:LeNettest}, using LeNet, it can be deduced that our method performs better than previously mentioned non-structured weight pruning techniques~\cite{Pruning1,trainspnn}. Also, the proposed structured global pruning method achieves comparable performance as~\cite{Pruning2}, without requiring customized hardware and software to realize the full advantage of the method's compression.

According to the experiments on different sized datasets with different model architectures, our
method shows superior compression performance without a significant loss of accuracy using VGG-16, ResNet-56 and
ResNet-110 on CIFAR and ResNet-34 on ImageNet. It is worth to mention that by retraining
for extra 20 epochs for AlexNet, the error reported is decreased by more than 0.2\%, which
could be done for all other networks. This may indicate that these results do not represent
the bottleneck compression results provided more training. Most importantly, we show
that by retraining ResNet-50 for fewer number of epochs, still a significant amount of
compression can be reached without a notable loss in accuracy, and by extended retraining, more compression can be achieved as well as a moderately compressed network with a slightly better accuracy than the base model. It is worth to mention that
although we only reported the models'~compression percentages achieved, in all networks
that contain the the majority of the number of parameters in the convolutional layers, the
compression percentages directly converts to reduction in computation needed by the model,
as the computational power of a neural networks model is dominated by the computational
requirements of convolutional layers.

\begin{figure*}[h!]
\begin{minipage}{\textwidth}
\centering
\begin{subfigure}[b]{0.475\textwidth}
\centering
   \includegraphics[width=1\linewidth]{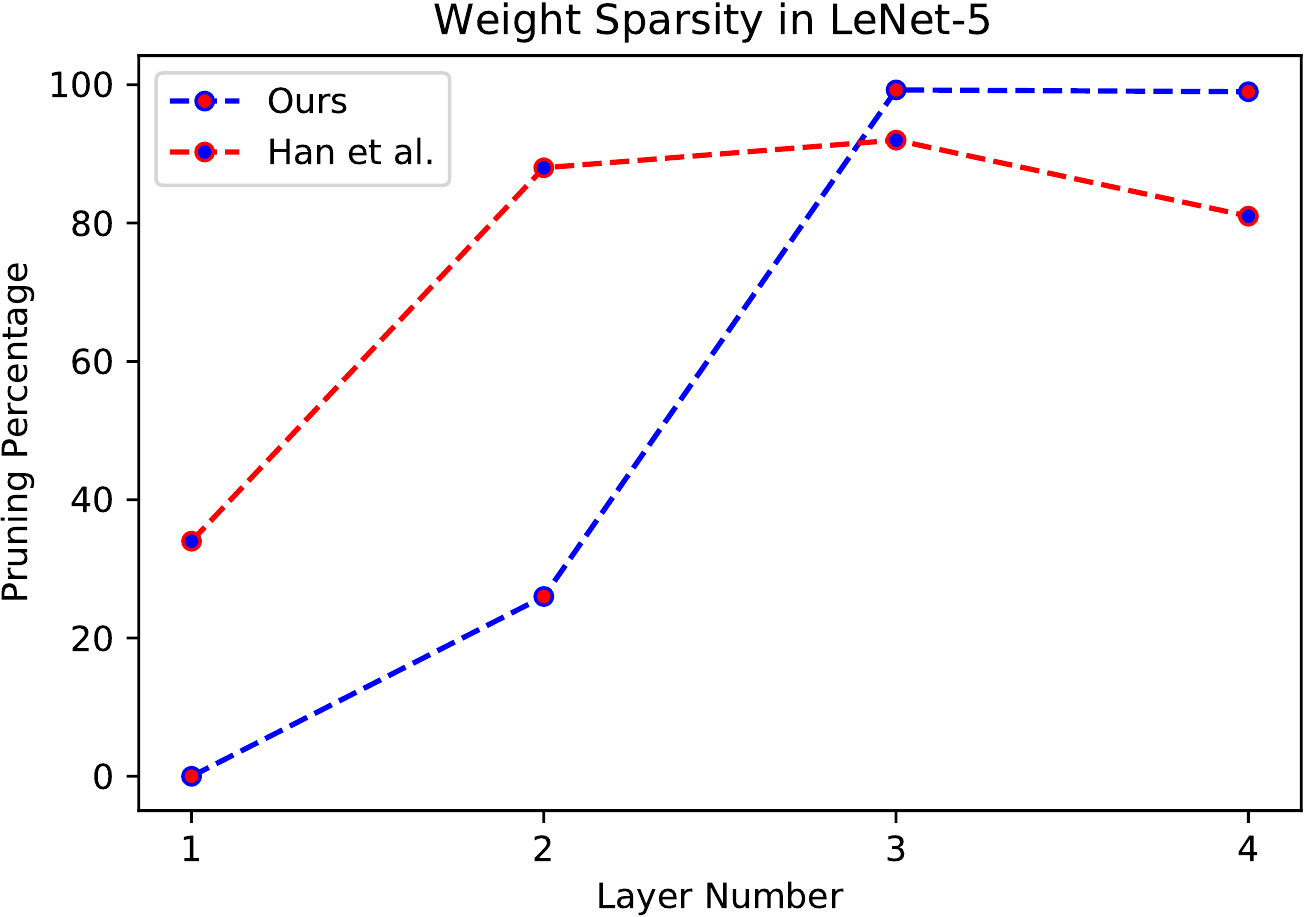}
\caption{LeNet-5 on MNIST}
\label{fig:weightssparsity}
\end{subfigure}
\bigskip 
\quad
 \begin{subfigure}[b]{0.475\textwidth}  
 \centering 
   \includegraphics[width=1\linewidth]{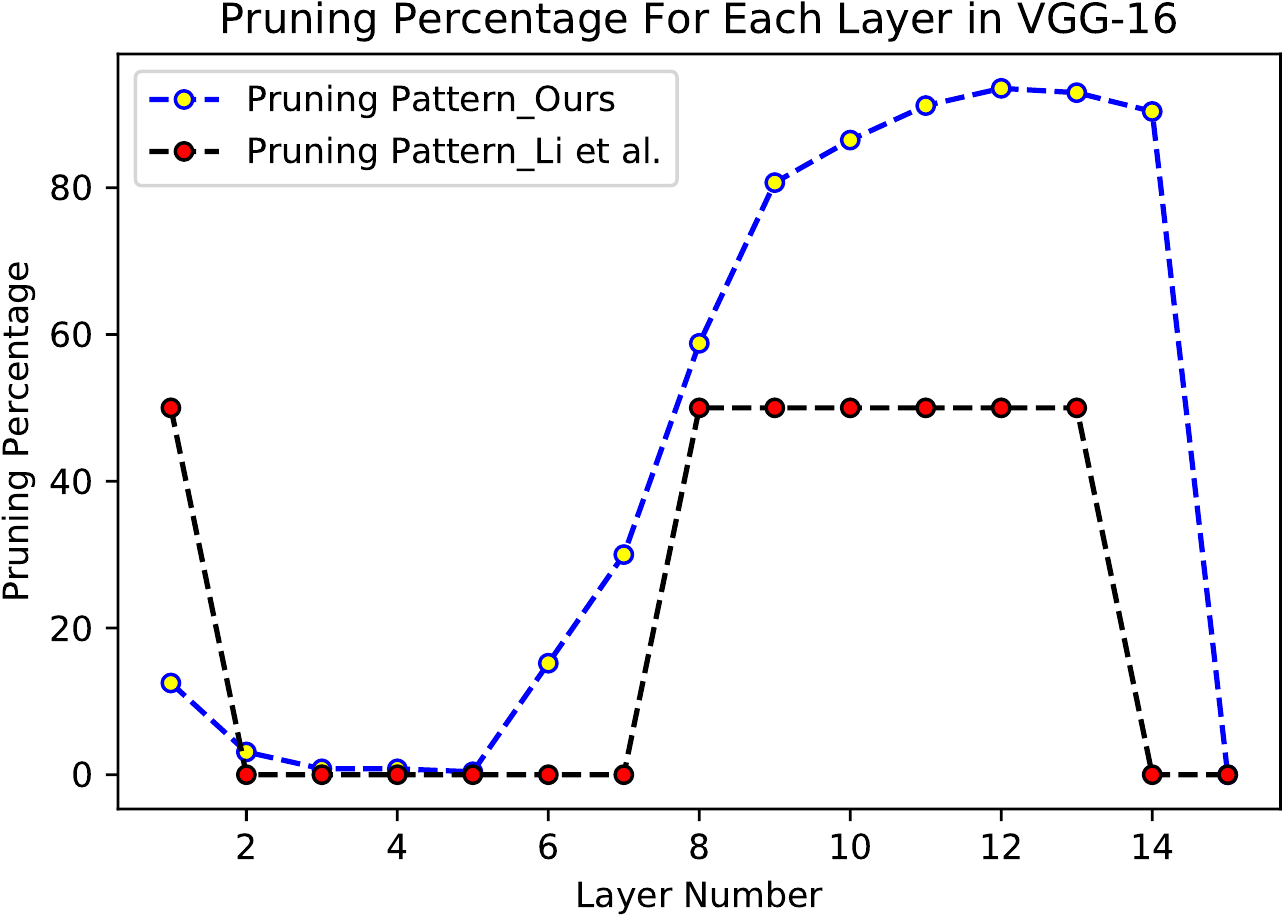}
\caption{VGG-16 on CIFAR10.}
\label{fig:patternVGG}
\end{subfigure}
\begin{subfigure}[b]{0.475\textwidth}
\centering
   \includegraphics[width=1\linewidth]{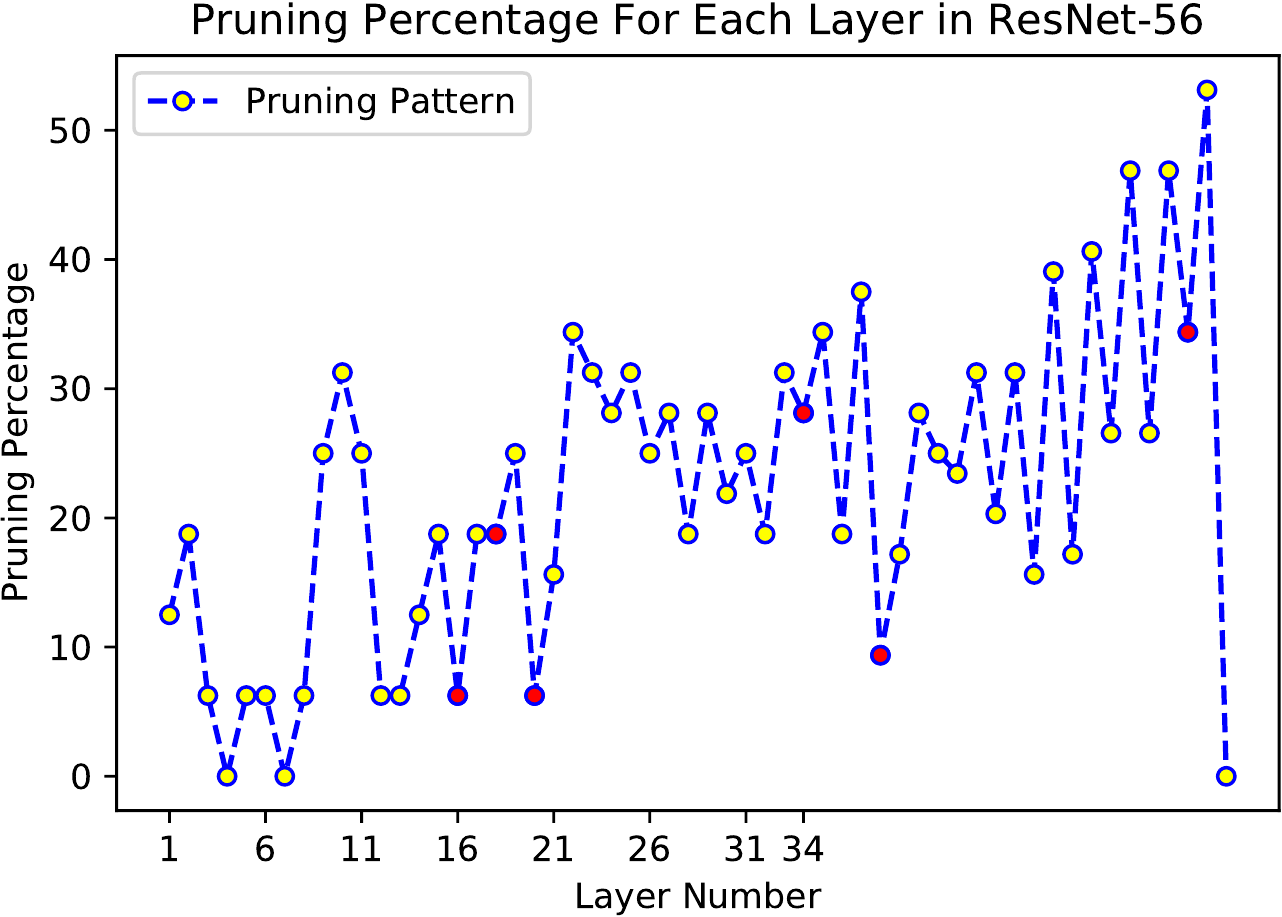}
\caption{ResNet-56 on CIFAR10.}
\label{fig:PruningResCI}
\end{subfigure}
\quad
\begin{subfigure}[b]{0.475\textwidth}   
 \centering 
   \includegraphics[width=1\linewidth]{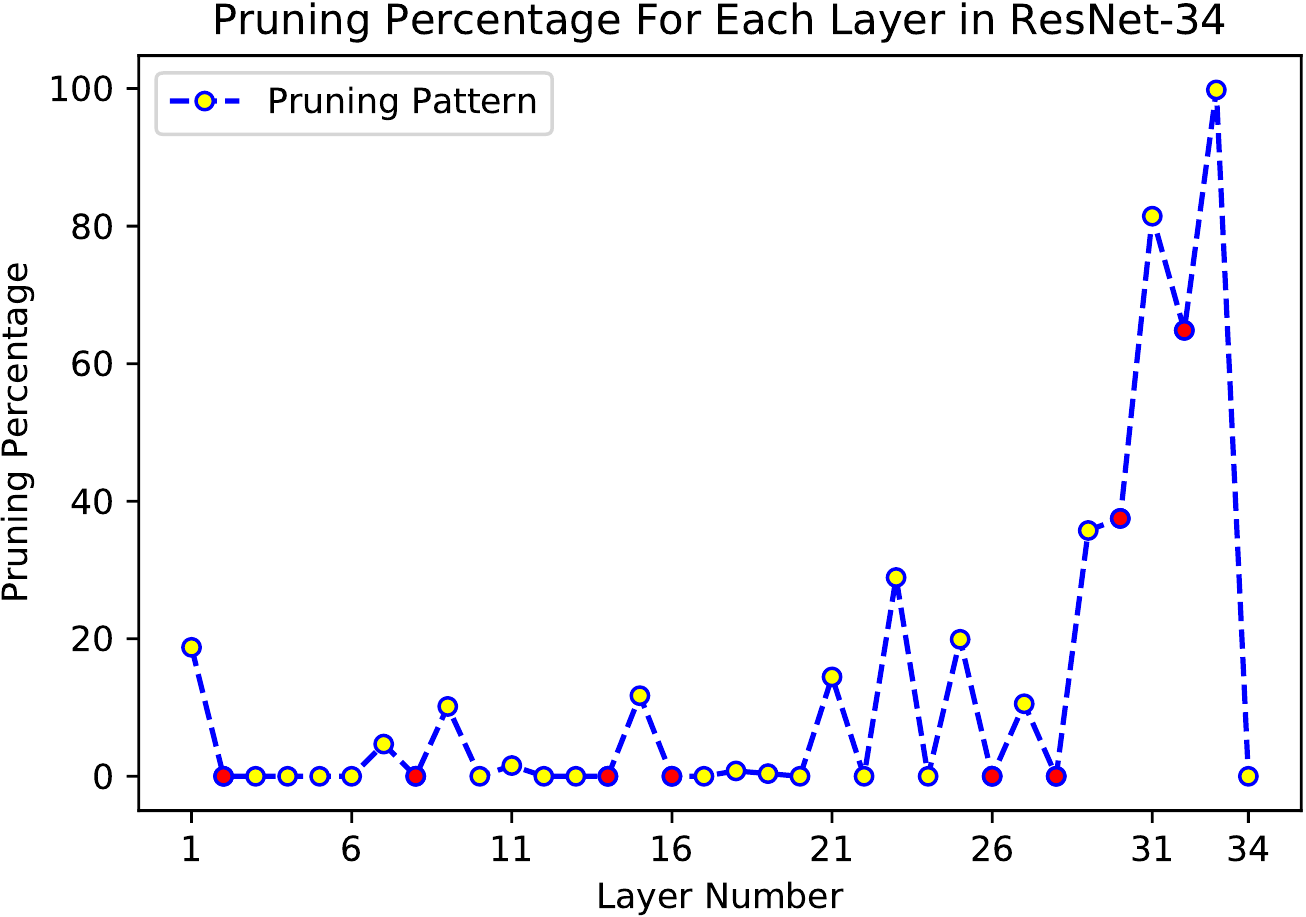}
\caption{ResNet-34 on ImageNet.}
\label{fig:ResNetim}
\end{subfigure}
\caption{Pruning Patterns: (a) Weights sparsity in each layer for LeNet-5 after pruning, compared for our method vs. Han et al.~\cite{Pruning1}. (b) Pruning Pattern of VGG-16 on CIFAR10 for our method vs. Li et al.~\cite{pruningfilters}. (c) Pruning Pattern of ResNet-56 on CIFAR10, the red marked points are the sensitive layers reported by~\cite{pruningfilters}. (d) Pruning Pattern of ResNet-34 on ImageNet, the red marked points are the sensitive layers reported by~\cite{pruningfilters}.}
\end{minipage}
\end{figure*}

\subsection{Global Pruning for Layer Importance Calculation}
In this part of the experiments, we explore the patterns of layer pruned and capture the behavior of global pruning on realizing the most sensitive layers. To be able to compare with the other structured pruning methods, we plot the pruning percentage of each layer according to percentage of filters/neurons pruned from the layer.

On MNIST, in Fig.~\ref{fig:weightssparsity}, a comparison between our method and~\cite{Pruning1} in terms of the resulted sparsity of the weights is performed since the comparison is done with a non-structured pruning method. Without any pre-calculation of layer importance, we observe the same weights' pruning pattern in our method, where the first layer is the least pruned layer and the third layer is the most pruned. It is worth mentioning that the percentage of pruning observed in the classification layer is due to the pruned neurons of its previous layer, thus pruning the input features to the layer.

As for CIFAR, in Fig.~\ref{fig:patternVGG}, we visualize the pruning pattern for VGG-16 and it can be observed that the deeper layers are generally more pruned. Moreover, also without any pre-calculation or manual tuning of layer pruning percentage, we observe a very similar pruning pattern of our method, compared to~\cite{pruningfilters}. It is also observed that our method lead to a much smoother curve as no layer is manually restricted from being pruned.

The pruning pattern on CIFAR for ResNet-56 is also shown on Fig.~\ref{fig:PruningResCI}, from which the deeper layers are shown to be more pruned, moreover, the sensitive layers reported by~\cite{pruningfilters} corresponds to local minima in terms of the pruning percentages, showing higher sensitivity to pruning with compared to surrounding layers. These sensitive layers include those that lie at residual blocks close to the layers where the number of feature maps changes.

We also experiment using ResNet-34 on ImageNet in Fig.~ \ref{fig:ResNetim}. The same trend as before is observed, as the deeper layers are more pruned than their shallower counterparts. Also similarly, the sensitive layers reported by~\cite{pruningfilters} corresponds to local minimum points in terms of compression in our method, confirming the previous observations on CIFAR.

\begin{table}[h!]
\begin{center}
\begin{tabular}{lccc}
\hline
Method   & Error\% & Param\% & Eff. Param\% \\ \hline
Baseline & 0.80      &   - &   -          \\
Non-Structured  & 0.77      & 93.04    & 86.08              \\
Non-Global    & 0.76      & 89.80    & 89.80              \\
Ours-Oneshot  & 0.80      & 96.06    & 96.06              \\
Ours  & \textbf{0.75}      & \textbf{97.40}   & \textbf{97.40}             \\
\hline
\end{tabular}
\end{center}
\caption{Evaluation of different components of proposed method. Param.\% is the parameters' pruning percentage; Eff.Param\%. is the effective parameters' pruning percentage with taking into account the extra indices storage for non-structured pruning as studied by~\cite{Sparsesaving}.} 
\label{table:Components}
\end{table}

The previous results show that despite of no direct pre-calculation of layer importance, our global pruning method proved the effectiveness of using global ranking of filter according to their L1-norm as a global importance parameter. This is proven by finding similar pruning patterns and sensitive layers to other methods which use heuristics to calculate layer importance before pruning to decide the pruning percentage for each layer~\cite{pruningfilters,Pruning1}. However, in contrast to~\cite{pruningfilters}, we find that earlier layers are less pruned than the deeper layers. We suggest that our evaluation of layer's sensitivity to pruning is more accurate because of our global importance score for each filter. The effectiveness of global pruning suggests that parameter importance is not only intra-layer dependent, but is also inter-layer dependent.

\subsection{Effect of Global Pruning on Different Parameters}
In this part of experiments, the effect of global pruning is tested. For the sake of simplicity we used LeNet-5 and MNIST for these tests.

\subsubsection{Evolution over pruning iterations}
\begin{figure}[h!]
\begin{center}                      
   \includegraphics[width=0.87\linewidth]{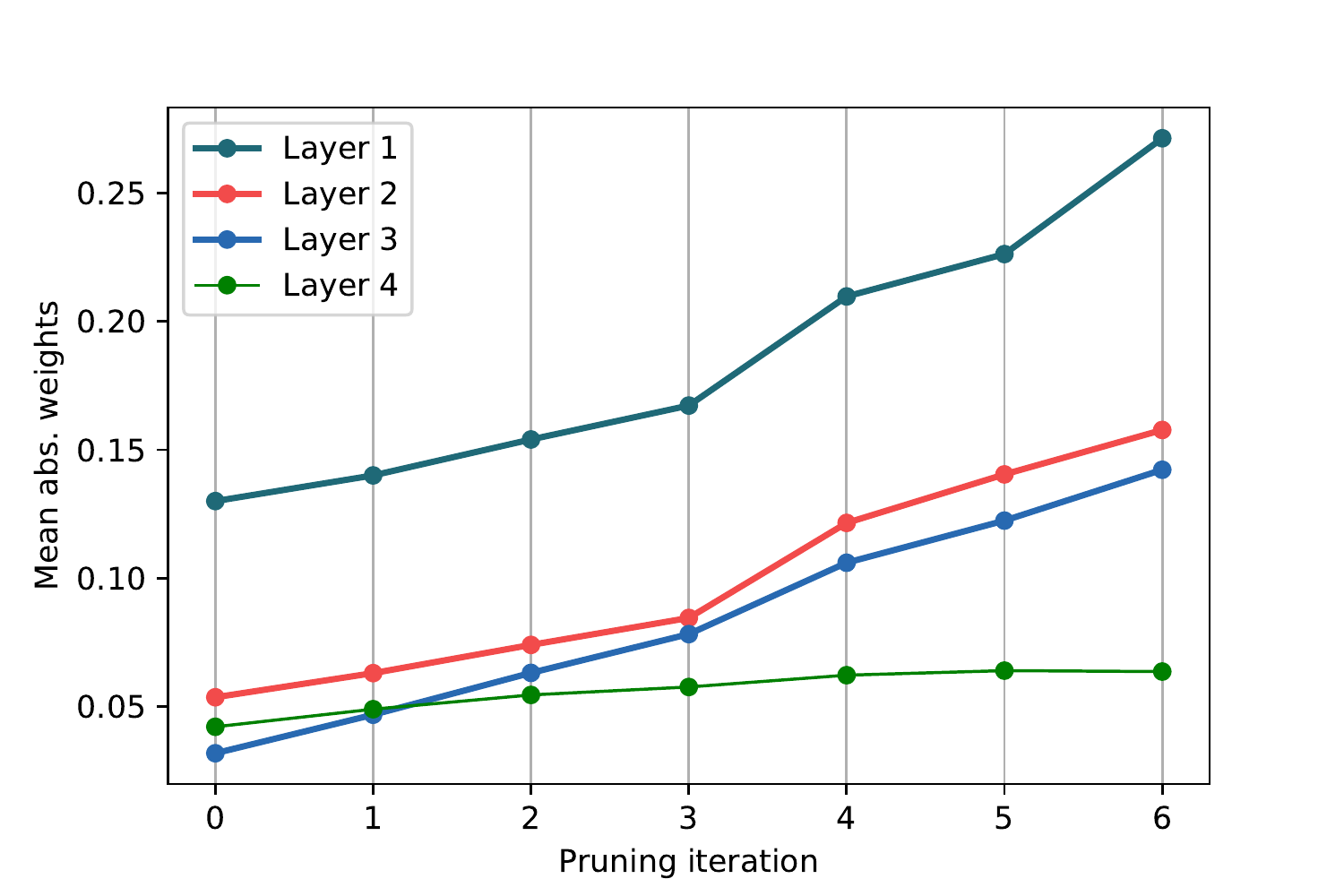}
\end{center}
   \caption{Mean absolute value of the weights for each layer dynamically over pruning iterations; Iteration 0 refer to the fully trained LeNet-5 model.}
\label{fig:weightsdist}
\end{figure}

It is fairly intuitive that the network gets more compressed every pruning iteration. But to understand the effect of the pruning on the remaining unpruned parameters, we observe the evolution of the remaining weights' distribution after each pruning iteration in Fig.~\ref{fig:weightsdist}. As observed, the weights' distribution is changed every iteration, and the remaining weights tend to have more magnitude as the network is more pruned, which suggests that the unpruned weights become more active to compensate for the pruned ones.

\subsubsection{Pruned Network Neuron Analysis}
To understand why the structurally pruned network can have the same baseline accuracy, and how the unpruned parameters compensate for their pruned counterparts. In addition to that why structured pruning is more promising than unstructured weight pruning. Fig.~\ref{fig:neuronsdist} shows the final number of neurons for each magnitude range after all pruning and retraining iterations are commenced. We matched the pruning of our method and its unstructured counter part to be at 94\%. It is obvious after retraining that our method has a higher percentage of very weak neurons (mostly pruned) and a higher percentage of strong neurons, while the non-structured version has the highest percentage of neurons having intermediate values.

The previous results suggest that method pushes the remaining neurons to be more active to compensate for the pruned neurons and is better in this regard than the non-structured weight pruning version.

\subsection{Structured Pruning Efficiency}
We conduct this experiment to explore the efficiency of structured pruning on changing the network's architecture. For that, we compare our method vs Han et al.~\cite{Pruning1} on the difference of the weights sparsity of both methods as well as neurons' sparsity for each layer in Fig.~ \ref{fig:weightssparsity}. Although Han et al.~\cite{Pruning1} weight pruning percentage is higher in terms of the first and second layer, the difference of the effective neuron sparsity in negligible, while on the third layer, the effective neuron sparsity is much higher in our case.

The reason why the high weight pruning percentage corresponds to high neuron percentage is that we target removing all weights leading to a filter/neuron, thus introducing similar percentage to neuron pruning to that of the weight pruning.

\subsection{Ablation Study}
Finally, we perform an ablation study by using one-shot pruning to test different components of our method; structured pruning and global pruning. Both components are analyzed by removing a component each test, resulting in: \textbf{i)} Non-Structured - pruning applied on weights separately. \textbf{ii)} Non-Global - every layer is pruned individually. Then, the effect of the pruning strategy on the method with all its components is analyzed by comparing: \textbf{i)} Ours-Oneshot - using one-shot pruning and \textbf{ii)} Ours - using iterative pruning. We compare using the error percentage, the pruning percentage and the effective pruning percentage with taking into account the extra indices storage for non-structured pruning as studied by \cite{Sparsesaving}.

By comparing the previous versions that use one-shot pruning, our method has fewer parameters with a negligible classification error increase; (Non-Structured and Non-Global). Also, applying pruning iteratively is superior to one-shot pruning (Tab. \ref{table:Components}). Pruning iteratively instead of one-shot, gives the parameters the opportunity to adapt to the performance loss. Moreover, global pruning and structured pruning has a significant role of increasing the compression results.  

\begin{figure}[]

  \centering
  \centerline{\includegraphics[width=0.9\linewidth]{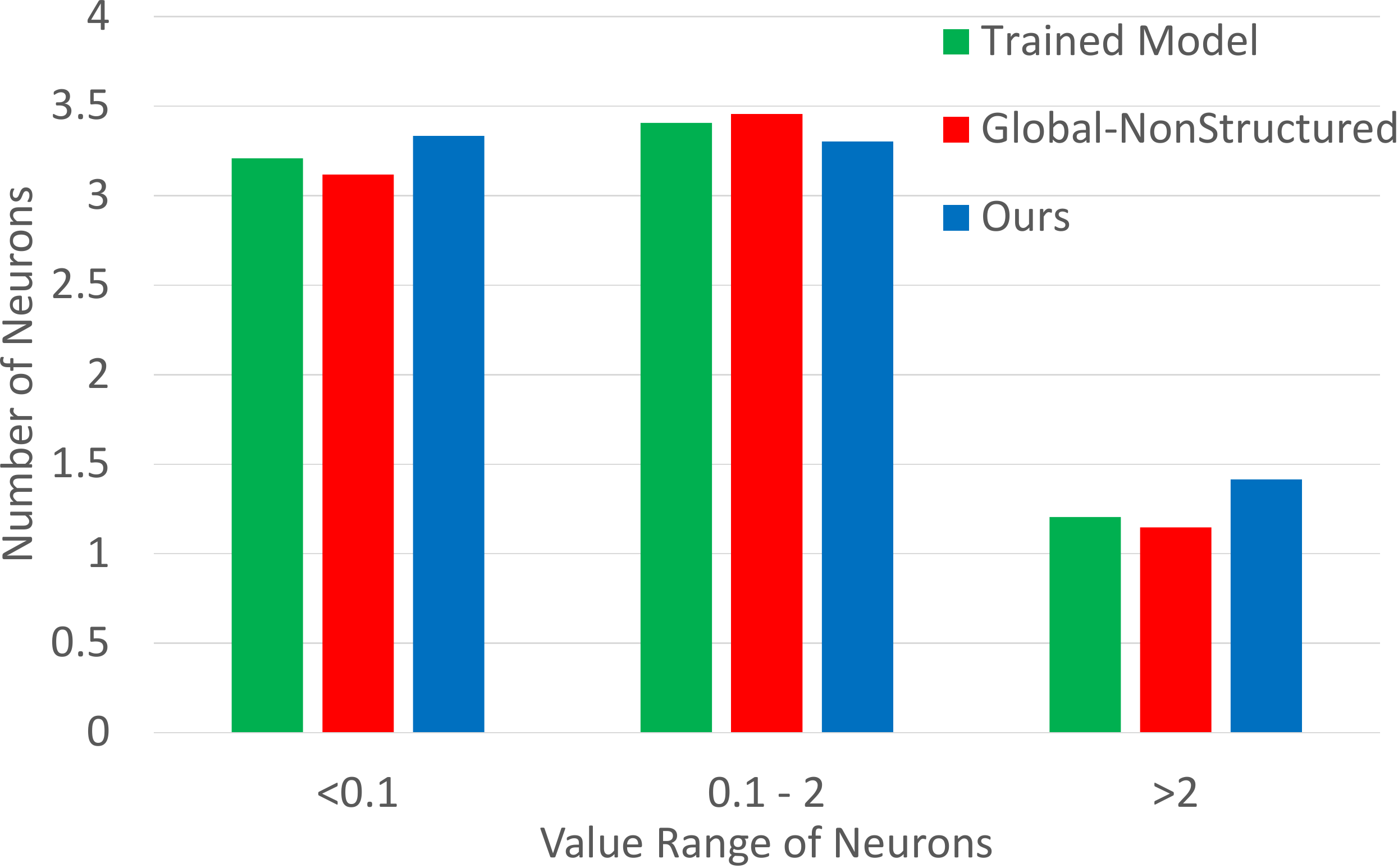}}

\caption{Effect of different methods on the neurons' distribution; the x-axis shows the magnitude value range of neurons, while the y-axis the number of neuron in log-scale.}
\label{fig:neuronsdist}
\end{figure}

Finally, it can be deduced from the results of Tab. \ref{fig:neuronsdist} that although our method prunes a lot of weak neurons, it encourages more activated neurons when compared to the unstructured counterpart, which can explain why our method exhibits higher accuracy in spite of higher compression rates (Tab. \ref{table:Components}).

\begin{figure}[h!]
\begin{center}
   \includegraphics[width=0.87\linewidth]{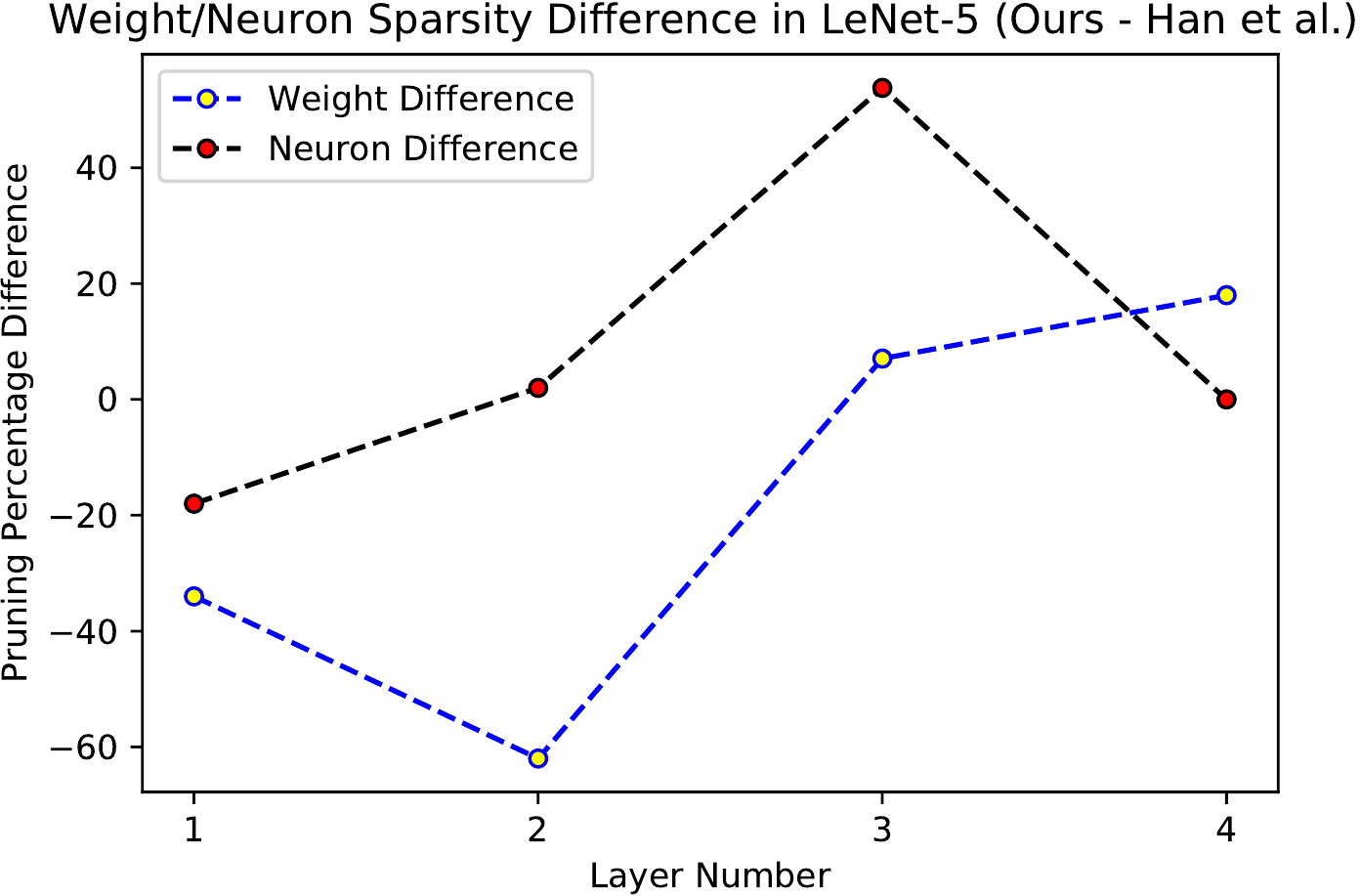}
\end{center}
   \caption{Weight and neuron sparsity difference in LeNet-5 between our method vs. Han et al.~\cite{Pruning1}.}
\label{fig:neuronssparsity}
\end{figure}

\section{Discussion}

Our method exhibits superior performance in terms of compression and error score as shown on the tests on VGG-16 and ResNet (Table \ref{table:Benchmark}) performed on CIFAR10. On ImageNet, our method out-performed other structured pruning methods mentioned, compressing AlexNet without a significant loss of accuracy, while producing a highly compressed ResNet-34 and ResNet-50 without a significant loss of accuracy and provide an extensively-retrained but moderately compressed ResNet-50 that slightly surpasses the base model in terms of accuracy.

Moreover, studying the components of the method in Table (\ref{table:Components}) shows that every component in the method has a significant impact on increasing compression. Moreover, the ablation study shows that the layers' pruning distribution of our method is similar to \cite{Pruning1} on LeNet-5 except filters/neurons pruned are significantly higher. Additionally, from the pruning patterns, it can be deduced that although our method has the same pattern as \cite{Pruning1}, significantly higher number of neurons was pruned. Finally, it can be deduced from the results of Table \ref{fig:neuronsdist} that although our method prunes a lot of weak neurons, it encourages more activated neurons when compared to the unstructured counterpart, which can explain why our method exhibits higher accuracy in spite of higher compression rates (Table \ref{table:Components}).
\section{Conclusion}
We presented a novel structured pruning method to compress neural networks without significantly losing accuracy. By pruning filters according to a global threshold based on ranking of all L1-norm of all filters across the network, we succeeded to deduce an efficient hidden filter importance score without any extra calculation. Our method was able to highly compress models on CIFAR-10 such as VGG-16, ResNet-56 and ResNet-110. Moreover, on ImageNet we reached higher compression percentages using AlexNet and ResNet-34. We presented a less-retrained ResNet-50 which introduced high compression performance and a extensively-retrained ResNet-50 that slightly surpassed the base model accuracy with 30\% less size. Also, on LeNet-5 we show that only 11K parameters are sufficient to exceed the baseline performance, compressing more than 97\%. Such compression performance can be attributed to accurate pruning patterns' calculation, and high accuracy can be attributed to a much higher number of activated neurons and weights after pruning and retraining. To realize the advantages of our method, no dedicated hardware or library are needed. For future work, we plan to present a version of the method which is optimized for decreasing computation. Moreover we are motivated to compress state-of-the-art DenseNet on ImageNet. Finally, we
plan to further investigate structured pruning and coupling it with an information theoretic view.

\vfill\pagebreak

% References should be produced using the bibtex program from suitable
% BiBTeX files (here: strings, refs, manuals). The IEEEbib.bst bibliography
% style file from IEEE produces unsorted bibliography list.
% -------------------------------------------------------------------------
\bibliographystyle{IEEEbib}
\bibliography{arxiv_paper}

\end{document}